\documentclass[12pt]{article}
\usepackage[margin=1in]{geometry}
\usepackage[font=small,labelfont=bf]{caption}
\usepackage{times}
\usepackage{graphicx}
\usepackage{subcaption}
\usepackage{colortbl}
\graphicspath{ {images/} }
\usepackage{amsmath}
\usepackage{times}
\usepackage{array}
\usepackage{multirow}% http://ctan.org/pkg/multirow
\usepackage{multicol}
\usepackage{tabularx}
\usepackage{booktabs}
\usepackage[table]{xcolor}
\usepackage[utf8]{inputenc}

\usepackage[round]{natbib}
\usepackage[english]{babel}
\usepackage{titlesec}
\titleformat{\section}
{\normalfont\fontsize{14}{15}\bfseries}{\thesection}{1em}{}
\titleformat{\subsection}
{\normalfont\fontsize{12}{15}\bfseries}{\thesubsection}{1em}{}

\title{\vspace{-2.0cm}%\rule{\linewidth}{1mm}	
	Dealing with Stochasticity in Biological ODE Models	%\rule{\linewidth}{0.5mm}
	}
\author{{\normalsize \textbf{Hamda bint e Ajmal, Michael Madden and Catherine Enright}}\\{\normalsize  Discipline of Information Technology, National University of Ireland Galway}\\
	{\small 	Email:\email{h.ajmal1@nuigalway.ie}, \email{michael.madden@nuigalway.ie}, \email{cathenright@gmail.com}} }
\date{}
\begin{document} 
	\maketitle
	\vspace{-1cm}
\begin{abstract}% \vspace{-2ex}
Mathematical modeling with Ordinary Differential Equations (ODEs) has proven to be extremely successful in a variety of fields, including biology. However, these models are completely deterministic given a certain set of initial conditions. We convert mathematical ODE models of three benchmark biological systems to Dynamic Bayesian Networks (DBNs). The DBN model can handle model uncertainty and data uncertainty in a principled manner. They can be used for temporal data mining for noisy and missing variables. We apply Particle Filtering algorithm to infer the model variables by re-estimating the models parameters of various biological ODE models. The model parameters are automatically re-estimated using temporal evidence in the form of data streams. The results show that DBNs are capable of inferring the model variables of the ODE model with high accuracy in situations where data is missing, incomplete, sparse and irregular and true values of model parameters are not known.
\end{abstract}

\section*{Introduction}
Mathematical models of physical systems are widely available in many domains \citep{ottesen2004applied}. These models embody existing expert knowledge and can be considered sufficient statistics of all prior experimentation in the domain. 

Mathematical modeling with ODEs has a very long tradition in biology. The earliest biological ODE models date back to 18th century \citep{malthus1798essay, verhulst1845mathematical} . ODE models assume that the observed dynamics of a system are exclusively driven by internal, deterministic mechanisms and there is no uncertainty in the process. However, in reality, biological systems are always subject to 'unexplainable' influences which are neither easily understood nor is it feasible to model them explicitly. All biological dynamical systems evolve under stochastic forces. Models of biological systems are concerned with subsystems of the real world, therefore ideally, they must include the effect of external random influences as they cannot be completely isolated from model \citep{ditlevsen2013introduction}. There are many factors in biological processes that are difficult to model explicitly with ODEs, for example, harmonic oscillations, enzymatic processes, or individual characteristics like body mass index, genes, lifestyle, habits, age etc. Such factors can contribute to the erratic behavior in a model. Other factors may include measurement noise or data uncertainty. These multiple sources of noise need to be modeled explicitly, however, simple ODE models tend to ignore them, which effects the analysis of the studied biological system and makes them impractical to be applied to the real world. There is also increasing evidence of stochastic behavior in critical biological processes, such as gene regulation and cellular behavior. All these factors point to the possible need to account for stochasticity in mathematical models \citep{lipniacki2006stochastic, lipniacki2006transcriptional}. 

ODE models generally describe population level behaviors. For example, ODE models already exist to describe the pharmacokinetics and pharmacodynamics of all drugs \citep[see][]{shargel2007applied}. They are developed during clinical trials, published for all approved drugs and used to determine safe dosage regimes for the average patient. Therefore, parameterization of these models is generic, typically done in a theoretical manner or based on laboratory or experimental data. However, the critically ill patient’s unique parameters can vary considerably from those of the average patient, such population based models often fail to capture specific clinical scenarios. 

To describe individuals, model parameters must be re-calibrated using observations of the individual. However, the observed data may be missing or noisy, or it could sparse or infrequent relative to the dynamics of the underlying system thus, making individualisation a challenging task.

Previous research work \citep{enright2015modelling} in our research group  has shown that we can overcome the limitations of ODE models by mapping them to Dynamic Bayesian Networks (DBNs) incorporating a first order Euler solver. DBNs are well suited to handle uncertainty and deal with noisy and missing data. Learning the structure of the DBNs is a challenging task, especially in situations where data is sparse or incomplete. The expert knowledge available in form of ODEs can be utilized as ODEs are encapsulated into the DBN structure. DBNs can reason efficiently with this powerful combination of domain knowledge and real time data. They explicitly model measurement uncertainty and parameter uncertainty, allowing model parameters to be adjusted from initial approximate values to their correct values using real-time evidence. By doing this, the knowledge elicitation bottleneck is bypassed. The technique was previously applied to the problem of modeling glycaemia in patients in an Intensive Care Unit \citep{enright2010}, producing promising results.

We apply the methodology of encapsulating ODE models into DBNs on three small to medium sized benchmark biological ODE models using the software package PROFET \footnote{http://profet.it.nuigalway.ie/} and evaluate the results of ODE model variable inference and model parameter re-estimation. 

\section{DBNs for Biological ODE Models}
In order to \emph{individualize} a general biological ODE model to a specific case, we convert it into a DBN using the software application PROFET. The DBN framework explicitly models noise as measurement and parameter uncertainty and then reduces the uncertainty over time by individualizing model parameters using temporal evidence. We apply this approach on three benchmark biological ODE models which were used by \citet{dondelinger2013ode} to apply their ODE model parameter inference methods. We evaluate the results of ODE model variable inference and model parameter re-estimation on these ODE models; however our experimental setup is different from that of \cite{dondelinger2013ode}. The essence of their work is to infer ODE model parameters from multiple noisy time series data. Our aim is to \emph{individualise} the model parameters on a single time series data stream. Therefore, for each ODE model, benchmark data is obtained using the R package deSolve. The values of the model parameters and the initial state of model variables to generate the benchmark data are taken from \citet{dondelinger2013ode}. Instead of adding noise to the data, we assume that the true values of the ODE model parameters are unknown and must be inferred from the population values. To discover the correct model parameters, evidence (which in these cases are the true values taken from the benchmark solutions) is incorporated. Complete details of methodology to convert ODE models to DBNs incorporating first order Euler solver can be found in \citet{enright2012thesis}. Below we describe each model and explain the results.
\subsection{The PIF4/5 Model}
As described in \citet{dondelinger2013ode}, this is a model of gene regulation of genes \textit{PIF4} and \textit{PIF5} by \textit{TOC1}  in the ciracadian clock gene regulatory network of \textit{Arabidopsis thaliana}. The complete network is described by a Locke 2-loop model \citep{locke2005}. Following \citeauthor{dondelinger2013ode} (\citeyear{dondelinger2013ode}), the model was simplified, such that, the genes \textit{PIF4} and \textit{PIF5} are combined as \textit{PIF4/5}.
% The regulation model is described by the following ODE:
%{\small \begin{align*}
%	\label{eq:6}
%	\tag{1}
%	\begin{split}
%	\frac{PIF4/5(t)}{dt} = s.\frac{K_d^h}{K_d^h + TOC1(t)^h} - d.PIF4/5(t)
%	\end{split}
%	\end{align*}}
The DBN created from the $PIF4/5$ ODE model is shown in Figure \ref{fig5:sfig1}. Model parameter $s$ is the promoter strength, $K_d$ is the rate constant, $h$ represents the Hill coefficient of the regulation by $TOC1$ and $d$ is the degradation rate of $PIF4/5$. $PIF4/5$ and $TOC1$ represent the concentration of the genes. The model parameters are modeled as continuous nodes with linear Gaussian distribution. The initial state model distribution of the DBN can be viewed as the distribution of the population. We assume that we do not know the real values of the model parameters but we only have a rough guess of the population parameters. The true values of the model parameters must be inferred from the data. Model parameters are allowed to vary in each time step. Evidence for the observed data is sampled from the benchmark data at different time points. The evidence is deliberately sampled at sparse and irregular intervals. However, it does not contain any noise. 

We run standard fixed time step Particle Filtering algorithm \citep{gordon1993novel} to infer the values of model parameters and model variables. Figure \ref{fig6:sfig1} shows a comparison of benchmark solution and the predicted solution. It can be seen that even though incorrect values of model parameters were chosen at the outset, the accuracy of the predicted values of $PIF4/5$ concentration begins to improve as evidence is incorporated into the system. The Root Mean Square Error (RMSE) and Mean Absolute Error (MAE) of the predicted data is 0.070 and 0.035 respectively.

\subsection{The Lotka-Volterra Model Predator Prey Model}
\begin{figure*}[h]
	%	\centering
		\begin{subfigure}[b]	{.25\textwidth}
			\centering
			\includegraphics[scale = 0.3]{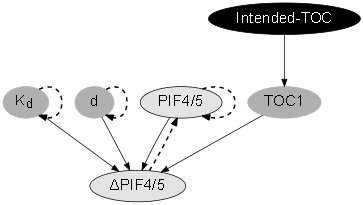}
			\subcaption{}
			\label{fig5:sfig1}
		\end{subfigure} 
	%	\hfil
	\begin{subfigure}[b]	{.25\textwidth}
		\centering
		\includegraphics[scale = 0.3]{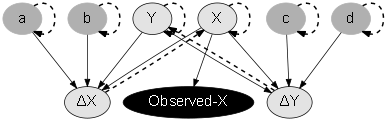}
		\subcaption{}
		\label{fig5:sfig2}
	\end{subfigure} 
	\hfil
		\begin{subfigure}[b]	{.40\textwidth}
			\centering
			\includegraphics[scale = 0.33]{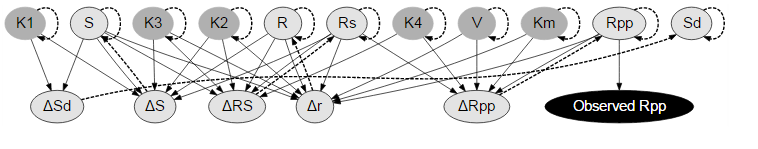}
			\subcaption{}
			\label{fig5:sfig3}
		\end{subfigure} 
	\caption{ (a) DBN structure of the (a) The PIF4/5 Model (b) The Lotka-Volterra model (c) The Signal Ttransduction Cascade Model} 
\end{figure*}

The Lotka-Volterra model, also known as the predator-prey model is described by two first order, non linear ODEs. These were first proposed by Edward J. Lotka in \citeyear{lotka1910} and since then have been frequently used to describe the dynamics of biological systems in which two species interact, one as the predator and the other as the prey. 
%{\small\begin{align*}
%	\label{eq:7}
%	\tag{7}
%	\begin{split}
%	\frac{dX(t)}{dt} &= a.X(t) - b.X(t).Y(t) 
%	\\
%	\frac{dY(t)}{dt} &= c.X(t)Y(t) - c.Y(t)
%	\end{split}
%	\end{align*}}
The DBN structure corresponding to the ODE model is shown in Figure \ref{fig5:sfig2}. The model variable node $X$ represent the number of prey and $Y$ represents the number of some predator. Model parameters $a,b,c$ and $d$ are the positive real numbers that describe the interaction of the two species. This ODE model exhibits periodicity and the interactions between model variables are non-linear.

We follow the similar steps described in the previous section and run the DBN inference in time interval [0,2]. Sparse evidence is sampled from the benchmark data of the model variable $X$. Graph in Figure \ref{fig6:sfig2} shows predicted values of prey population over time. As before, can see that even though incorrect values of model parameter were chosen at the start, the predicted results are very close to those of benchmark solution. RMSE and MAE of predicted values of $X$ is 0.287 and 0.137 respectively. 

\subsection{The Signal Transduction Cascade Model}
The model of signal transduction cascade was described in \citet{vyshemirsky2008bayesian}.
% using the following coupled ODEs. 
%{\small \begin{align*}
%	\label{eq:8}	\tag{8}
%	\begin{split}
%	\frac{dS(t)}{dt} &= -k_1.S(t) - k_2.S(t).R(t) + k_3.RS(t)
%	\\
%	\frac{dS_d(t)}{dt} &= k_1.S(t)
%	\\
%	\frac{dR(t)}{dt} &= -k_2.S(t).R(t) + k_3.RS(t) + \frac{V.Rpp(t)}{Km + Rpp(t)}
%	\\
%	\frac{dRS(t) }{dt} &= k_2.S(t).R(t) - k_3.RS(t) - k_4.RS(t)
%	\\
%	\frac{dRpp(t)}{dt} &= k_4.RS(t) - \frac{V4.Rpp(t)}{Km + Rpp(t)} 
%	\end{split}
%	\end{align*}}
The DBN constructed by PROFET for this ODE model is shown in Figure \ref{fig5:sfig3}. The input signal is represented by the concentration of protein $S$ which can bind to protein $R$ to form a complex $RS$ which activates protein $R$ into its phosphorylated form $Rpp$. Protein $Rpp$ can then be de-activated back to protein $R$. The model also defines the degradation of input signal, that is, the conversion of protein $S$ into its degraded form $dS$. 
This system represents a realistic formulation of signal transduction as a mathematical model using mass action and Michaelis-Menten kinetics.

\begin{figure*}[h]
	\centering
	\begin{subfigure}[b]	{.30\textwidth}
			\centering
			\includegraphics[width=\textwidth]{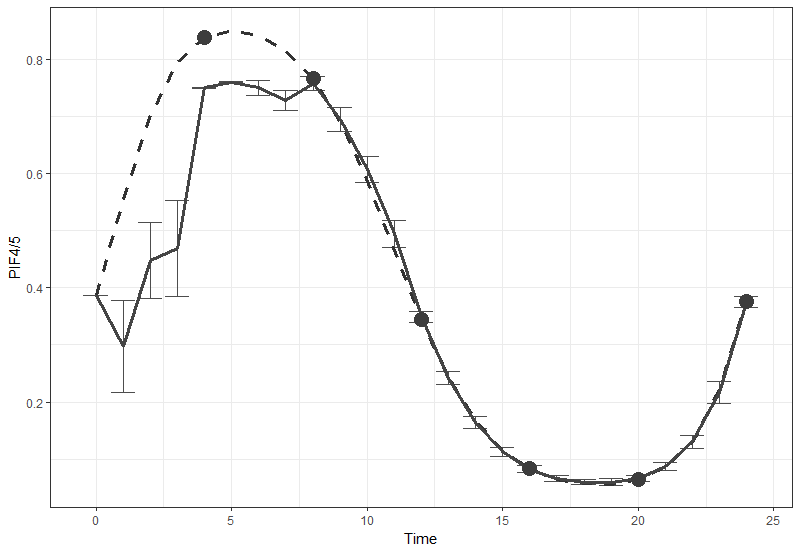}
			\subcaption{}
			\label{fig6:sfig1}
		\end{subfigure}
			\begin{subfigure}[b]	{.30\textwidth}
		\centering
		\includegraphics[width=\textwidth]{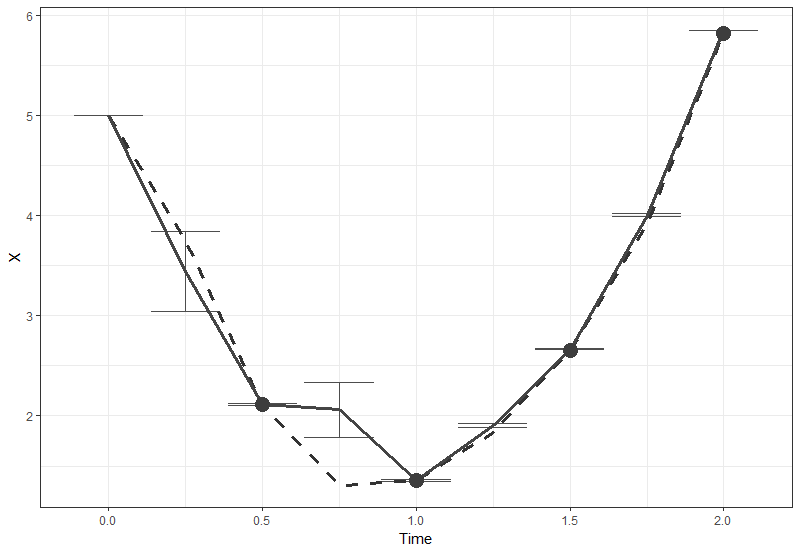}
		\subcaption{}
		\label{fig6:sfig2}
	\end{subfigure}
	\begin{subfigure}[b]	{.30\textwidth}
		\centering
		\includegraphics[width=\textwidth]{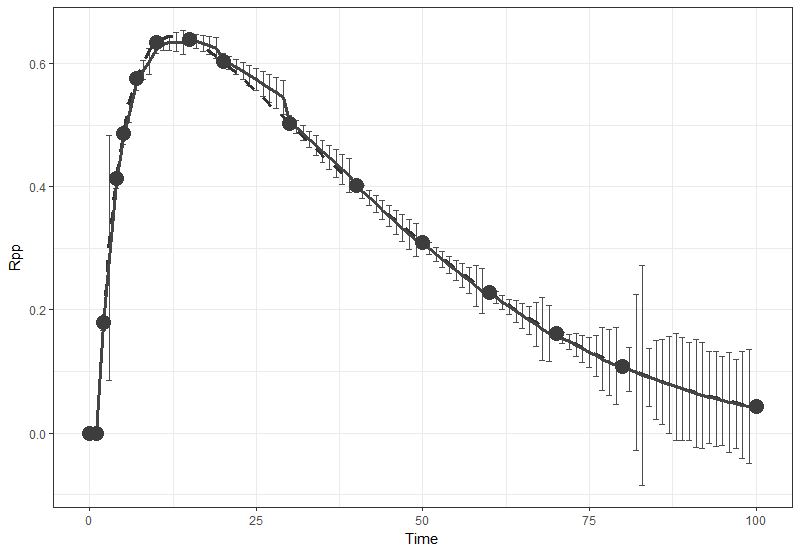}
		\subcaption{}
		\label{fig6:sfig3}
	\end{subfigure}
	\caption{Predicted values of model variables in the ODE system. Dashed lines represent the benchmark solution. Solid lines are the predicted trajectories. Error bars show one standard deviation. Gray dots are the points where evidence is received. (a) The PIF4/5 model (b) Lotka-Volterra Model (c) The Signal Transduction Cascade model} 
\end{figure*}

We run the DBN inference in time interval [0,100]. Following \citet{dondelinger2013ode}, we sample the evidence of $Rpp$ concentration at more time points during the earlier part of the time series where the dynamics tend to be faster. In Figure \ref{fig6:sfig3}, we plot the value of $Rpp$ concentration predicted along with the benchmark solution. We can see that the DBN was able to predict the values of model variable $Rpp$ with high accuracy, even though incorrect model parameters were chosen in the outset. RMSE and MAE of the predicted data is 0.0085 and 0.0053 respectively.

\section*{Conclusions}
This paper has discussed the challenges of applying ODE modeling approaches to real world biological systems. ODE models are deterministic and they can not account for the uncertainty in the real world. We proposed to tackle this problem by converting the ODE models to DBNs that incorporate a first order Euler solver as proposed by \citet{enright2013}. We tested the methodology of various biological ODE models. We simulated a real-life situation by assuming that the true values of the ODE model parameters are not known and data is sparse, incomplete and collected at irregular intervals. Our DBN framework uses Particle Filtering inference algorithm to infer the values of model variables of the ODE model by re-estimating the values of model parameters at each time step. We have shown that our DBN modeling and inference works well for this task.

{\small \bibliography{scibib}}

\begin{thebibliography}{16}
\providecommand{\natexlab}[1]{#1}
\providecommand{\url}[1]{\texttt{#1}}
\expandafter\ifx\csname urlstyle\endcsname\relax
  \providecommand{\doi}[1]{doi: #1}\else
  \providecommand{\doi}{doi: \begingroup \urlstyle{rm}\Url}\fi

\bibitem[Ditlevsen and Samson(2013)]{ditlevsen2013introduction}
Susanne Ditlevsen and Adeline Samson.
\newblock Introduction to stochastic models in biology.
\newblock In \emph{Stochastic biomathematical models}, pages 3--35. Springer,
  2013.

\bibitem[Dondelinger et~al.(2013)Dondelinger, Husmeier, Rogers, and
  Filippone]{dondelinger2013ode}
Frank Dondelinger, Dirk Husmeier, Simon Rogers, and Maurizio Filippone.
\newblock {ODE parameter inference using adaptive gradient matching with
  Gaussian processes}.
\newblock In \emph{AISTATS}, 2013.

\bibitem[Enright(2012)]{enright2012thesis}
Catherine Enright.
\newblock \emph{A probabilistic framework based on mathematical models with
  application to medical data streams}.
\newblock PhD thesis, 2012.

\bibitem[Enright et~al.(2010)Enright, Madden, Stuart~Russell, Manley, Laffey,
  Harte, Mulvey, and Madden]{enright2010}
Catherine Enright, Michael~G Madden, Norm~Aleks Stuart~Russell, Geoffrey
  Manley, John Laffey, Brian Harte, Anne Mulvey, and Niall Madden.
\newblock Modelling glycaemia in {ICU} patients a dynamic {B}ayesian network
  approach.
\newblock In \emph{Proceedings of BIOSIGNALS-2010, 3rd International Joint
  Conference on Biomedical Engineering Systems and Technologies}, Valencia,
  2010.

\bibitem[Enright and Madden(2015)]{enright2015modelling}
Catherine~G Enright and Michael~G Madden.
\newblock Modelling and monitoring the individual patient in real time.
\newblock In P.F.~Lucas A.~Hommersom, editor, \emph{Foundations of Biomedical
  Knowledge Representation}, pages 107--136. Springer, 2015.

\bibitem[Enright et~al.(2013)Enright, Madden, and Madden]{enright2013}
Catherine~G Enright, Michael~G Madden, and Niall Madden.
\newblock Bayesian networks for mathematical models: techniques for automatic
  construction and efficient inference.
\newblock \emph{International Journal of Approximate Reasoning}, 54\penalty0
  (2), 2013.

\bibitem[Gordon et~al.(1993)Gordon, Salmond, and Smith]{gordon1993novel}
Neil~J Gordon, David~J Salmond, and Adrian~FM Smith.
\newblock {Novel approach to nonlinear/non-Gaussian Bayesian state estimation}.
\newblock In \emph{IEE Proceedings F-Radar and Signal Processing}, volume 140.
  IET, 1993.

\bibitem[Lipniacki et~al.(2006{\natexlab{a}})Lipniacki, Paszek, Brasier, Luxon,
  and Kimmel]{lipniacki2006stochastic}
Tomasz Lipniacki, Pawel Paszek, Allan~R Brasier, Bruce~A Luxon, and Marek
  Kimmel.
\newblock Stochastic regulation in early immune response.
\newblock \emph{Biophysical journal}, 90:\penalty0 725--742,
  2006{\natexlab{a}}.

\bibitem[Lipniacki et~al.(2006{\natexlab{b}})Lipniacki, Paszek,
  Marciniak-Czochra, Brasier, and Kimmel]{lipniacki2006transcriptional}
Tomasz Lipniacki, Pawel Paszek, Anna Marciniak-Czochra, Allan~R Brasier, and
  Marek Kimmel.
\newblock Transcriptional stochasticity in gene expression.
\newblock \emph{Journal of theoretical biology}, 238\penalty0 (2):\penalty0
  348--367, 2006{\natexlab{b}}.

\bibitem[Locke et~al.(2005)Locke, Southern, Kozma-Bogn{\'a}r, Hibberd, Brown,
  Turner, and Millar]{locke2005}
James~CW Locke, Megan~M Southern, L{\'a}szl{\'o} Kozma-Bogn{\'a}r, Victoria
  Hibberd, Paul~E Brown, Matthew~S Turner, and Andrew~J Millar.
\newblock Extension of a genetic network model by iterative experimentation and
  mathematical analysis.
\newblock \emph{Molecular systems biology}, 1\penalty0 (1), 2005.

\bibitem[Lotka(1910)]{lotka1910}
Alfred~J Lotka.
\newblock Contribution to the theory of periodic reactions.
\newblock \emph{The Journal of Physical Chemistry}, 14\penalty0 (3):\penalty0
  271--274, 1910.

\bibitem[Malthus(1798)]{malthus1798essay}
Thomas Malthus.
\newblock An essay on the principle of population.
\newblock 1798.

\bibitem[Ottesen et~al.(2004)Ottesen, Olufsen, and Larsen]{ottesen2004applied}
Johnny~T Ottesen, Mette~S Olufsen, and Jesper~K Larsen.
\newblock \emph{Applied mathematical models in human physiology}.
\newblock Siam, 2004.

\bibitem[Shargel et~al.(2007)Shargel, Wu-Pong, and Yu]{shargel2007applied}
Leon Shargel, Susanna Wu-Pong, and Andrew~BC Yu.
\newblock \emph{{Applied biopharmaceutics \& pharmacokinetics}}.
\newblock McGraw-Hill, 2007.

\bibitem[Verhulst(1845)]{verhulst1845mathematical}
PF~Verhulst.
\newblock Mathematical researches into the law of population growth increase.
\newblock \emph{Nouveaux M{\'e}moires de l’Acad{\'e}mie Royale des Sciences
  et Belles-Lettres de Bruxelles}, 18:\penalty0 1--42, 1845.

\bibitem[Vyshemirsky and Girolami(2008)]{vyshemirsky2008bayesian}
Vladislav Vyshemirsky and Mark~A Girolami.
\newblock Bayesian ranking of biochemical system models.
\newblock \emph{Bioinformatics}, 24\penalty0 (6), 2008.

\end{thebibliography}

\clearpage

\end{document}